\documentclass{article}

\usepackage[preprint]{neurips_2025}
\workshoptitle{DL4CODE}

\usepackage[utf8]{inputenc} 
\usepackage[T1]{fontenc}    
\usepackage{hyperref}       
\usepackage{url}            
\usepackage{booktabs}       
\usepackage{amsfonts}       
\usepackage{nicefrac}       
\usepackage{microtype}      
\usepackage{xcolor}         
\usepackage{amsmath}
\usepackage{multirow}
\usepackage{subcaption}
\usepackage{graphicx}
\usepackage{tabularx}
\usepackage{array}

\newcommand{\model}{\href{https://huggingface.co/jinaai/jina-reranker-v3}{\texttt{jina-reranker-v3}}}
\newcommand{\docembtoken}{\textcolor{cyan}{\texttt{<|doc\_emb|>}}}
\newcommand{\queryembtoken}{\textcolor{cyan}{\texttt{<|query\_emb|>}}}

\newcommand{\jinaembeddings}{\texttt{jina-embeddings-v3}}
\newcommand{\jinarerankerv}{\texttt{jina-reranker-v2}}
\newcommand{\jinarerankerme}{\texttt{jina-reranker-m0}}
\newcommand{\jinacolbertv}{\texttt{jina-colbert-v2}}
\newcommand{\bgereranker}{\texttt{bge-reranker-v2-m3}}
\newcommand{\mxbaibase}{\texttt{mxbai-rerank-base-v2}}
\newcommand{\mxbailarge}{\texttt{mxbai-rerank-large-v2}}
\newcommand{\qwenbase}{\texttt{Qwen3-0.6B}}
\newcommand{\qwenrerankersmall}{\texttt{Qwen3-Reranker-0.6B}}
\newcommand{\qwenrankerlarge}{\texttt{Qwen3-Reranker-4B}}

\newcommand{\jinacodeemb}{\texttt{jina-code-embeddings-0.5b}}

\title{\model{}: Last but Not Late Interaction for Listwise Document Reranking}

\author{%
  Feng Wang\textsuperscript{1} \quad Yuqing Li\textsuperscript{1,2} \quad Han Xiao\textsuperscript{1} \\
  \\
  \textsuperscript{1}Jina AI GmbH \quad
  \textsuperscript{2}University of Pittsburgh \\
  Prinzessinnenstraße 19, 10969, Berlin, Germany \\
  \texttt{research@jina.ai} \\
}
\begin{document}

\maketitle

\begin{abstract}
\model{} is a 0.6B-parameter multilingual listwise reranker that introduces a novel \emph{last but not late} interaction. Unlike late interaction models like ColBERT that encode documents separately before multi-vector matching, our approach applies causal attention between the query and all candidate documents in the same context window, enabling rich interactions before extracting contextual embeddings from each document's final token. The new model achieves state-of-the-art BEIR performance with 61.85 nDCG@10 while being significantly smaller than other models with comparable performance.
\end{abstract}

\section{Introduction}

Neural document retrieval faces a fundamental efficiency-effectiveness tradeoff. Cross-encoders achieve strong performance through joint query-document processing but require separate forward passes for each pair, while embedding models enable efficient similarity computation but lose fine-grained interaction signals. Recent models have attempted to bridge this gap through different interaction approaches. Late interaction models like ColBERT~\citep{khattab2020colbert} and their variants~\citep{liu2024analysis,jha2024jina} separately encode queries and documents into multi-vector representations, then perform interaction through token-level similarity operations.

We introduce \model{}, which features a novel \emph{last but not late} interaction (LBNL) that takes a fundamentally different approach from existing methods. While late interaction models delay attention until after encoding documents separately, our method applies causal attention between the query and all documents within the context window, enabling cross-document interactions before extracting contextual embeddings from each document's \textbf{\emph{last}} token. Unlike late interaction models that interact after encoding, we enable interactions during encoding---making our approach \textbf{\emph{not late}}. This ``listwise'' processing is not possible with separate encoding or bi-encoder approaches and represents our core innovation.

Evaluation shows \model{} achieves 61.85 nDCG@10 on BEIR~\citep{thakur2021beir}, representing the highest score among all evaluated rerankers and a 4.79\% improvement over our previous \jinarerankerv{}. The model excels particularly in multi-hop retrieval with HotpotQA reaching 78.58, fact verification achieving 94.01 on FEVER, competitive multilingual performance across 18 languages at 66.83 on MIRACL~\citep{miracl2023} and crosslingual retrieval with 67.92 Recall@10 on MKQA~\citep{mkqa2020} across 26 languages, and code retrieval reaching 70.64 on CoIR~\citep{li2024coir}.

\section{Related Work}

Document reranking approaches can be categorized by their interactions and learning objectives. Traditional learning-to-rank methods~\citep{bruch2023efficient} include pointwise approaches that predict relevance scores independently, pairwise methods like RankNet~\citep{burges2005learning} that compare document pairs, and listwise techniques that optimize global ranking objectives. Cross-encoders like BERT-based rerankers~\citep{nogueira2019passage} achieve strong performance through full query-document interaction but require separate forward passes for each pair, creating computational bottlenecks for large-scale retrieval. Recent comparative studies~\citep{dejean2024thorough} demonstrate that while LLM-based rerankers show impressive zero-shot capabilities, traditional cross-encoders remain highly competitive across diverse retrieval scenarios.

Late interaction models represent a significant approach that balances efficiency with expressiveness. ColBERT~\citep{khattab2020colbert} exemplifies this approach by independently encoding queries and documents into multi-vector representations, then computing similarity through MaxSim operations over token-level embeddings. This design enables pre-computation of document representations while preserving fine-grained matching signals. Recent developments have expanded this approach: analysis of matching mechanisms and token pruning strategies~\citep{liu2024analysis} provides theoretical foundations, LITE~\citep{ji2024efficient} introduces learnable late interactions, and Jina-ColBERT-v2~\citep{jha2024jina} extends the approach to multilingual settings. PyLate~\citep{chaffin2025pylate} provides flexible frameworks for training and deployment of such models. The late chunking method~\citep{gunther2024late} processes complete documents through transformers before applying chunking boundaries, extracting chunk-level embeddings that preserve contextual relationships. This approach demonstrates how leveraging broader document context can improve embedding quality, though it focuses primarily on retrieval rather than reranking applications.

LLM-powered reranker has emerged as a powerful family with diverse implementations. These approaches can be categorized into discriminative and generative methods. Generative approaches like RankGPT~\citep{qin2023large} prompt LLMs to generate ranked lists, leveraging their reasoning capabilities for relevance assessment, but typically require large models for competitive performance. Fine-tuning methods like RankVicuna~\citep{pradeep2023rankvicuna} adapt existing models for relevance scoring tasks. Efficiency-focused innovations include FIRST~\citep{wu2024first}, which accelerates inference through single-token decoding, and PE-Rank~\citep{qin2024leveraging}, which leverages passage embeddings to reduce computational latency by $4.5\times$. Recent advances in training methodology include ERank~\citep{cai2025erank}, which combines supervised fine-tuning with reinforcement learning for improved ranking quality, and the Qwen3 Embedding series~\citep{yang2025qwen3embedding}, which demonstrates sophisticated multi-stage training pipelines. DeAR~\citep{abdallah2025dear} introduces dual-stage reasoning with LLM distillation for enhanced cross-document analysis.

\section{Model Architecture}

\model{} implements a new interaction that fundamentally differs from existing approaches. Built upon Qwen3-0.6B~\citep{yang2025qwen3} with 28 transformer layers, 1024 hidden dimensions, 16 attention heads, and 131K token context capacity, our approach processes queries and multiple documents simultaneously within shared context windows. We add a lightweight MLP projector (1024$\rightarrow$512$\rightarrow$512 dimensions) to transform contextual representations into ranking-optimized embeddings. Table~\ref{tab:model_config} provides complete architectural specifications.

\subsection{Architecture}

\begin{figure}[!htbp]
\centering
\includegraphics[width=0.95\textwidth]{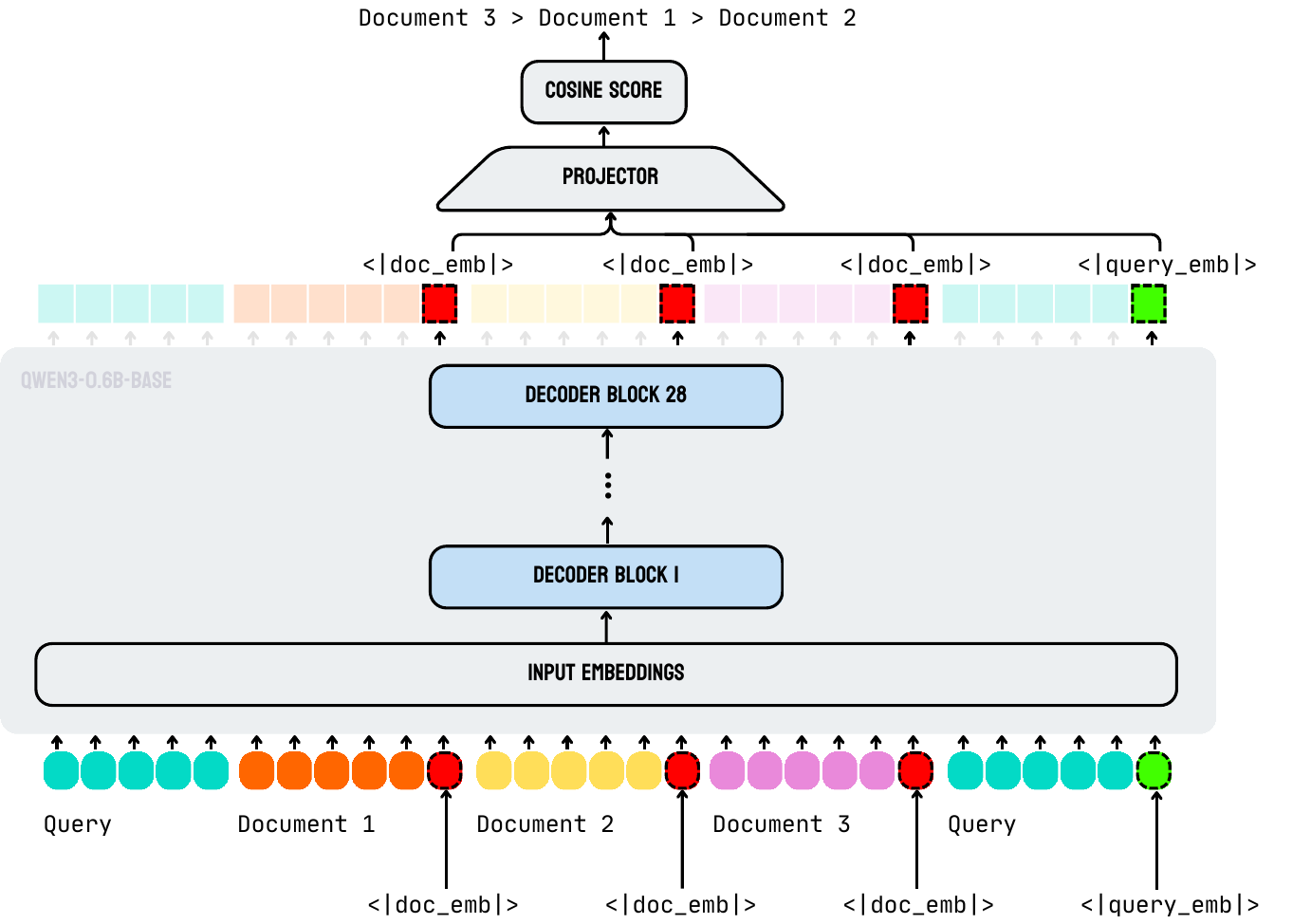}
\caption{Architecture of \model{} showing the transformer backbone with special token positions for embedding extraction. The model processes multiple documents and query in one context window, extracting contextual embeddings at designated token positions for similarity computation.}
\label{fig:architecture}
\end{figure}

Figure~\ref{fig:architecture} illustrates our architecture that addresses fundamental limitations in existing interactions. ColBERT~\citep{khattab2020colbert} achieves efficiency through separate encoding followed by multi-vector interaction, but cannot capture early query-document interactions during encoding or enable cross-document interactions within the attention mechanism. 

Our LBNL approach enables causal self-attention interaction within the transformer architecture: instead of delaying interaction until after separate encoding as in late interaction models, we process all documents and the query simultaneously within shared context windows. This allows each document to attend to other documents and observe their content, enabling contextual embeddings that capture not just query-document relevance but also inter-document relationships and comparative context. Such cross-document interactions are impossible in separate encoding approaches and represent a fundamental advancement in \model{} architecture.

We extract contextual embeddings at designated special token positions: $\tilde{\mathbf{q}} = \mathbf{H}_{t_q}$ and $\tilde{\mathbf{d}}_i = \mathbf{H}_{t_i}$ where $t_q$ and $t_i$ are positions of the special tokens and $\mathbf{H}$ represents the transformer's final layer hidden states after causal self-attention. These embeddings capture both local document semantics and global cross-document context through the shared attention mechanism, enabling rich inter-document interactions unavailable in separate encoding approaches.

A two-layer projection network with ReLU activation maps the 1024-dimensional hidden states to 256-dimensional embedding space: $\mathbf{q} = P_\phi(\tilde{\mathbf{q}})$ and $\mathbf{d}_i = P_\phi(\tilde{\mathbf{d}}_i)$. Relevance scores are computed via cosine similarity: $s_i = \cos(\mathbf{q}, \mathbf{d}_i)$. This architecture combines the expressiveness of joint encoding with efficient similarity computation.

For document collections exceeding the 131K token context limit, we process documents in batches of up to 64 documents per forward pass, with query embeddings maintained consistently across batches to ensure ranking coherence.

\subsection{Prompt Template}

\model{} processes structured prompts following Qwen3's instruction format with \texttt{system}/\texttt{user}/\texttt{assistant} roles to leverage existing instruction-following capabilities. As shown in Table~\ref{tab:prompt_template}, the system prompt establishes a search relevance expert persona, while the user prompt provides clear ranking instructions with dual query placement.

\begin{table}[!htbp]
\centering
\small
\begin{tabular}{l}
\toprule
\textbf{Prompt Template} \\
\midrule
\texttt{<|im\_start|>system} \\
\texttt{You are a search relevance expert who can determine} \\
\texttt{a ranking of passages based on their relevance to the query.} \\
\texttt{<|im\_end|>} \\
\hline
\texttt{<|im\_start|>user} \\
\texttt{I will provide you with k passages, each indicated by a numerical identifier.} \\
\texttt{Rank the passages based on their relevance to query: \textcolor{blue}{\textbf{[QUERY]}}} \\
\\
\texttt{<passage id="1">} \\
\texttt{\textcolor{red}{\textbf{[DOCUMENT\_1]}}\textbf{\docembtoken}} \\
\texttt{</passage>} \\
\texttt{<passage id="2">} \\
\texttt{\textcolor{red}{\textbf{[DOCUMENT\_2]}}\textbf{\docembtoken}} \\
\texttt{</passage>} \\
\texttt{...} \\
\texttt{<passage id="k">} \\
\texttt{\textcolor{red}{\textbf{[DOCUMENT\_k]}}\textbf{\docembtoken}} \\
\texttt{</passage>} \\
\\
\texttt{<query>} \\
\texttt{\textcolor{blue}{\textbf{[QUERY]}}\textbf{\queryembtoken}} \\
\texttt{</query>} \\
\texttt{<|im\_end|>} \\
\bottomrule
\end{tabular}
\caption{Complete prompt template structure used by \model{}. Special tokens \textbf{\docembtoken{}} and \textbf{\queryembtoken{}} mark positions for embedding extraction from transformer hidden states. 
}
\label{tab:prompt_template}
\end{table}

The template strategically places the query both at the beginning for instructions and at the end for final attention, sandwiching all documents in between. This design enables the final query position to attend to all preceding documents through causal attention while maintaining clear task instructions. Special tokens \docembtoken{} after each document and \queryembtoken{} after the final query mark specific positions for embedding extraction from transformer hidden states. 

\section{Training}

\subsection{Loss Functions}
\model{} employs a comprehensive multi-objective training approach combining InfoNCE loss with specialized auxiliary losses to optimize ranking performance across diverse domains.

The core training objective integrates multiple loss components, each addressing distinct aspects of the ranking problem:

\begin{equation}
\ell = \ell_{\mathrm{rank}} + 0.45 \cdot \ell_{\mathrm{disperse}} + 0.85 \cdot \ell_{\mathrm{dual}} + 0.85 \cdot \ell_{\mathrm{similar}}
\label{eq:overall_loss}
\end{equation}

The primary component is the InfoNCE loss $\ell_{\mathrm{rank}}$~\citep{oord2019representation}, which generates the core ranking signal through contrastive learning with hard negatives:
\begin{equation}
    \ell_{\mathrm{rank}} = - \frac{1}{N}\sum_{i=1}^N \log \frac{e^{s(\mathbf{q}_i, \mathbf{d}_{i}^+) / \tau}}{Z_i} \quad \text{where} \quad Z_i = e^{s(\mathbf{q}_i, \mathbf{d}_i^+)/\tau} + \sum_{k=1}^K e^{s(\mathbf{q}_i, \mathbf{d}_{i,k}^-)/\tau}
\label{eq:rank_loss}
\end{equation}
Here, $\mathbf{q}_i$ denotes the query embedding, $\mathbf{d}_i^+$ represents the positive document embedding, $\mathbf{d}_{i,k}^-$ denotes one of $K$ negative document embeddings, $s(\cdot, \cdot)$ is the cosine similarity function, $\tau$ is the temperature parameter, and $N$ is the batch size.

To prevent representation collapse, we incorporate the dispersive loss $\ell_{\mathrm{disperse}}$~\citep{wang2024contrastive}, which enhances embedding diversity by maximizing the average pairwise cosine distance between document embeddings:
\begin{equation}
    \ell_{\mathrm{disperse}} = \frac{1}{N} \sum_{i=1}^N \log \frac{1}{K} \sum_{k=1}^K \left(e^{s(\mathbf{d}_i^+, \mathbf{d}_{i,k}^-) / \tau} + \sum_{k'=k}^{K-1} e^{s(\mathbf{d}_{i,k}^-, \mathbf{d}_{i,k'+1}^-) / \tau}\right)
\label{eq:disperse_loss}
\end{equation}

The dual matching loss $\ell_{\mathrm{dual}}$\footnote{During training, the special token \textbf{\queryembtoken{}} is inserted at the end of the query at the beginning of the input sequence.} follows the same formulation as Eq.~\ref{eq:rank_loss} but computes the query embedding from the query tokens at the sequence start. This enforces bidirectional consistency between query-to-document and document-to-query similarity scores, enhancing ranking robustness.

Finally, the similarity loss $\ell_{\mathrm{similar}}$~\citep{huang2024cosent} maintains semantic coherence at the document level. For each document in the input set, we create an augmented duplicate $\mathbf{d}_{i}^{*}$ through text augmentation techniques. The loss then treats the original document and its augmented version as a positive pair, while other documents serve as negatives. This encourages consistent embedding representations for semantically equivalent documents, even when their surface forms differ due to augmentation.

\subsection{Multi-Stage Training}

The training methodology follows a progressive three-stage approach designed for systematic complexity scaling:

\textbf{Stage 1: Foundation Specialization.} Starting from pretrained \qwenbase{}, we simultaneously train domain-specific configurations using LoRA fine-tuning with $r$=16 and $\alpha$=32 targeting all attention and FFN layers while freezing the backbone. The model processes training sequences containing 16 documents per query (one positive and 15 negative examples), with each document truncated or padded to 768 tokens, yielding a maximal total sequence length of 12,288 tokens. Training data is drawn from diverse datasets including BGE-M3~\citep{bge2024m3} for multilingual coverage across 15 languages, Cornstack~\citep{suresh2025cornstack} for code retrieval, as well as specialized datasets for biomedical~\citep{xu2024bmretriever} and instruction following~\citep{weller2024followir} configurations.

\textbf{Stage 2: Context and Hard Negative Mining.} This stage combines context extension and comprehensive robustness optimization. Context extension is implemented in two ways: (1) extending individual document length to 8,192 tokens through datasets like MLDR~\citep{bge2024m3} for long-document understanding, and (2) increasing the number of negative documents from 15 to 45 per query while maintaining the total sequence length under 131K tokens. Simultaneously, cross-system hard negative mining ensures robustness through specialized optimizations including \textit{jina-en-v2} for English performance, \textit{miracl-v2} for multilingual retrieval, \textit{cornstack-v2} for code understanding, and \textit{context-chunk-v3} for long-document processing. Training systematically mines hard negatives across multiple retrieval systems including BGE, Jina, GTE, and E5-Large with up to 25 negatives per query and very low temperature of 0.05, using key datasets including MS-MARCO~\cite{nguyen2016msmarco}, mMARCO~\cite{bonifacio2021mmarco}, and domain-specific synthetic question-answer pairs.

\textbf{Stage 3: Model Ensemble and Optimization.} The final stage combines multiple specialized models trained in previous stages through linear model merging. Each domain-specific model contributes weighted expertise, with merge weights ranging from 0.25 to 0.65 based on domain importance and performance. This approach enables the final model to leverage diverse domain knowledge while maintaining architectural efficiency.

Detailed hyperparameter evolution across stages demonstrates multi-objective optimization with stage-tailored configurations (see Appendix~\ref{tab:training_config}). Foundation stages use aggressive learning rates of 5e-5 with substantial negative sampling of 15 negatives. Context scaling stages reduce batch sizes dramatically from 60 to 6 to accommodate 8K sequences while employing conservative learning rates of 6e-6. Loss weight adaptation varies across different domain specializations, with dispersive loss typically set to 0.45, dual-matching loss ranging from 0.65 to 0.85, and similarity loss stabilizing around 0.75 to 0.85 depending on the specific domain requirements.

\section{Evaluation}

\subsection{Experimental Setup}

Our evaluation spans four challenging benchmarks that test different aspects of ranking capability. BEIR~\cite{thakur2021beir} represents the gold standard for English retrieval evaluation, encompassing 13 heterogeneous tasks from question answering on Natural Questions to fact verification on FEVER, testing the model's ability to generalize across domains without task-specific optimization. MIRACL~\cite{zhang2022miracl} pushes multilingual boundaries with 18 languages spanning diverse linguistic families, from Arabic and Chinese to Finnish and Thai, requiring deep cross-lingual understanding. MKQA~\cite{longpre2021mkqa} specifically challenges cross-lingual question answering capabilities, while CoIR~\cite{li2023towards} focuses on the specialized domain of code retrieval, where semantic understanding of programming constructs becomes crucial.

The first-stage dense retriever is \jinaembeddings{}, providing the foundation top-100 candidates that all rerankers process. Second-stage rerankers encompass our previous \jinarerankerv{}, the multilingual \bgereranker{}, the mxbai-rerank variants at different scales, and \qwenrerankersmall{} and \qwenrankerlarge{} models. 

\subsection{Overall Performance Across Benchmarks}

Table~\ref{tab:overall_eval_result} demonstrates \model{}'s exceptional performance density across diverse evaluation scenarios. On BEIR, our model achieves the highest score among all rerankers at 61.85, establishing new state-of-the-art performance for English retrieval. This represents a 4.79\% improvement over our previous \jinarerankerv{} at 57.06, directly attributable to LBNL interaction mechanism where query and document embeddings are extracted from shared forward passes rather than separate encoding pipelines.

Parameter efficiency analysis reveals striking advantages compared to larger alternatives. Against the 1.5B parameter \mxbailarge{}, \model{} achieves superior BEIR performance with 61.85 versus 61.44 using $2.5\times$ fewer parameters, while providing specialized domain coverage unavailable in competing models reaching 70.64 on CoIR. This efficiency derives from architectural innovations: Qwen3's optimized transformer backbone combined with our specialized 512-dimensional projector network that concentrates ranking signals without requiring massive parameter scaling.

Multilingual evaluation reveals strong cross-lingual capabilities despite the model's compact architecture. The 66.83 score on MIRACL, while 2.49 points below the multilingual-specialized \bgereranker{} at 69.32, demonstrates effective knowledge transfer from our progressive training methodology. The 67.92 MKQA performance closely approaches \jinarerankerme{}'s 68.19, indicating that architectural sophistication can partially offset parameter differences in multilingual scenarios.

\begin{table}[!htbp]
\centering
\begin{tabularx}{\textwidth}{l c *{4}{>{\centering\arraybackslash}X}}
\toprule
\textbf{Models} & \textbf{\# Param} & \textbf{BEIR} & \textbf{MIRACL} & \textbf{MKQA} & \textbf{CoIR} \\
\midrule
\multicolumn{6}{c}{\textit{First-stage Retriever}} \\
\midrule
\jinaembeddings{} & 0.5B & 55.81 & 58.90 & 65.63 & -  \\
\jinacodeemb{} &-  &- &- &- & \textbf{73.94} \\
\midrule
\multicolumn{6}{c}{\textit{Second-stage Reranker}} \\
\midrule
\model{} & 0.6B & \textbf{61.85} & 66.83 & 67.92 & 70.64 \\
\jinarerankerv{} & 0.3B & 57.06 & 63.65 & 67.90 & 58.35 \\
\jinarerankerme{} & 2.4B & 58.95 & 66.75 & 68.19 & 66.89 \\
\bgereranker{} & 0.6B & 56.51 & \textbf{69.32} & 67.88 & 36.28 \\
\mxbaibase{} & 0.5B & 58.40 & 55.32 & 64.24 & 65.71 \\
\mxbailarge{} & 1.5B & 61.44 & 57.94 & 67.06 & 70.87 \\
\qwenrerankersmall{} & 0.6B & 56.28 & 57.70 & 65.34 & 65.18 \\
\qwenrankerlarge{} & 4.0B & 61.16 & 67.52 & \textbf{69.25} & 73.91 \\
\bottomrule
\end{tabularx}
\caption{Evaluation results for all rerankers. All scores are from our runs based on the top-100 retrieval results from the first row. For MKQA, we used Recall@10; for all other benchmarks, we used NDCG@10.}
\label{tab:overall_eval_result}
\end{table}

\subsection{English Retrieval Performance on BEIR}

Table~\ref{tab:beir_eval_result} provides granular analysis across BEIR's heterogeneous tasks, revealing specific architectural advantages. The model achieves consistent excellence across diverse reasoning tasks, with particularly strong performance on complex multi-hop reasoning reaching 78.58 on HotpotQA and fact verification achieving 94.01 on FEVER. These results highlight how LBNL interaction enables sophisticated query-document self-attention during encoding, capturing evidence relationships that separate encoding approaches miss.

Within the same scale category, \model{} reveals significant advantages. Against \bgereranker{} with the same 0.6B parameters, \model{} delivers a substantial 5.34\% improvement from 56.51 to 61.85, demonstrating architectural innovation over simple parameter scaling. The specialized 512-dimensional projector network effectively concentrates ranking signals while preserving contextual representations from the Qwen3 backbone. Remarkably, our model surpasses \mxbailarge{}'s 61.44 performance while using $2.5\times$ fewer parameters, establishing that sophisticated architecture can surpass brute-force scaling approaches.

Since \model{} processes all documents simultaneously in a listwise manner within shared context windows, we investigate the sensitivity to document ordering. We evaluate three variants: documents ordered by descending relevance scores (D), ascending scores (A), and random permutation (R). The results show modest variations across orderings, with random ordering (R) achieving the highest average of 62.24, followed by descending (D) at 61.85 and ascending (A) at 61.45. While the differences are not conclusive, this analysis reveals that the LBNL interaction maintains relatively stable performance across different input orderings, suggesting robust self-attention mechanisms that can effectively process documents regardless of their initial arrangement.

The model's dominance extends particularly to question-answering scenarios, where Natural Questions achieves 74.28 and argumentative retrieval on ArguAna reaches 73.43, showcasing the benefit of contextual embeddings. These tasks require understanding complex query intent and matching it against nuanced document semantics, precisely the scenario where our dual embedding extraction approach provides maximum advantage over traditional cross-encoder scoring.
\begin{table}[!htbp]
\centering
\scriptsize 
\begin{tabularx}{\textwidth}{l c c *{13}{>{\centering\arraybackslash}X}}
\toprule
\textbf{Models} & \textbf{Size} & \textbf{Avg.} & \textbf{TC} & \textbf{NFC} & \textbf{NQ} & \textbf{HQA} & \textbf{FQA} & \textbf{AA} & \textbf{TCH} & \textbf{DBP} & \textbf{SD} & \textbf{FVR} & \textbf{CFV} & \textbf{SF} & \textbf{QRA} \\
\midrule
\multicolumn{16}{c}{\textit{First-stage Retriever}} \\
\midrule
\jinaembeddings{} & 0.5B & 55.81 &	77.81 &	36.65 & 64.31 &	64.63 & 47.47 & 54.31 & 26.55 & 41.07 & 19.91 & 89.00 & 42.33 & 72.4 & 89.06 \\
\midrule
\multicolumn{16}{c}{\textit{Second-stage Reranker}} \\
\midrule
\model{} (D) & 0.6B & 61.85 & 84.75 & 37.66 & \textbf{74.28} & \textbf{78.58} & 49.16 & 73.43 & 32.24 &	47.98 &	23.23 &	\textbf{94.01} & 41.63 & 76.51 & \textbf{90.63}\\
\model{} (A) & 0.6B & 61.45 & 85.90 & 39.14 & 72.34 & 77.48 & 50.99 & 69.36 &	29.73 &	48.30 &	23.90 &	93.46 &	41.72 &	76.75 &	89.73\\
\model{} (R) & 0.6B & \textbf{62.24} & 86.59 & 38.92 & 72.90 & 78.03 & 51.81 & 74.12 &	30.12 &	48.37 &	24.26 &	93.84 &	43.05 &	76.84 & 90.24\\
\jinarerankerme{} & 2.4B & 58.95 & 84.17 & 41.03 & 72.25 & 76.99 & 51.62 & 40.69 & 31.79 & 49.34 & 22.91 & 91.14 & 36.42 & \textbf{79.94} & 88.01\\
\jinacolbertv{} & 0.6B & 54.49 & 81.94 & 35.88 & 66.01 & 74.36 & 43.62 & 35.46 & 29.11 & 47.14 & 19.40 & 87.92 & 29.20 & 70.13 & 88.25\\
\jinarerankerv{} & 0.3B & 57.06 & 80.53 & 37.17 & 67.39 & 76.17 & 46.48 & 39.28 &	32.35 &	47.81 &	20.03 &	93.02 &	37.17 &	76.50 & 87.83 \\
\bgereranker{} & 0.6B & 56.51 &	82.19 & 34.33 &	69.52 &	77.89 &	45.45 & 36.21 &	33.12 &	46.72 &	17.79 &	91.03 &	38.69 &	72.64 &	89.10 \\
\mxbaibase{} & 0.5B & 58.40 & 82.75	& 37.57 & 67.74 & 77.35 & 47.33 & 47.33 & 30.71 & 48.00 & 18.09 & 93.30 & 42.93 & 77.76 & 88.33 \\
\mxbailarge{} & 1.5B & 61.44 & 81.51 & 37.76 & 72.46 & 78.10 & \textbf{52.75} & \textbf{74.55} & 29.81 &	49.07 &	18.58 &	93.94 &	42.03 &	78.86 & 89.36 \\
\qwenrerankersmall{} & 0.6B & 56.28 & 87.08 & 38.37 & 56.54 & 74.41 & 43.45 & 56.53	& 27.26 & 43.54 & 20.98 & 86.19 & 44.11 & 74.89 & 78.32\\
\qwenrankerlarge{} & 4.0B & 61.16 & \textbf{87.08} & \textbf{41.56} & 69.06 & 77.03 & 52.29 & 58.82 &\textbf{33.73} & \textbf{50.81} & \textbf{26.01} & 87.80 &	\textbf{47.59} & 78.41 & 84.83 \\
\bottomrule
\end{tabularx}
\caption{Performances of different rerankers (nDCG@10 in \%) on BEIR. Top-100 retrieval results from \textbf{\jinaembeddings{}} are passed as input. The best results are marked in bold. Avg. represents the averaged result of the 13 BEIR datasets. For \model{}, (D)/(A)/(R) denote document ordering variants: Descending, Ascending, and Random relevance score ordering, respectively.}
\label{tab:beir_eval_result}
\end{table}

\subsection{Multilingual Performance on MIRACL}

MIRACL evaluation across 18 diverse languages demonstrates \model{}'s cross-lingual consistency despite its compact architecture. The 66.50 average performance reveals sophisticated multilingual understanding, with particularly strong results in morphologically complex languages like Arabic achieving 78.69 and challenging contexts like Thai reaching 81.06. These results reflect the effectiveness of our progressive multilingual training strategy, where architectural advantages help compensate for reduced multilingual specialization.

Perhaps most significantly, \model{} exhibits minimal performance degradation across linguistic families, from Indo-European languages like Russian at 65.20 to Sino-Tibetan languages like Thai at 81.06. This consistency stems from our progressive multilingual training strategy that incorporates diverse datasets including MIRACL, mMARCO, and domain-specific multilingual corpora during the three-stage training progression. The architectural advantage becomes particularly evident in Korean achieving 73.83, where the model's LBNL interaction enables effective handling of complex agglutinative morphology that traditional cross-encoders struggle to process efficiently.

Compared to \bgereranker{}'s dedicated multilingual optimization averaging 69.32, \model{} accepts a 2.82-point gap while achieving superior English performance and maintaining architectural efficiency. This trade-off reflects our design philosophy: contextual embedding extraction provides competitive multilingual capabilities without massive multilingual scaling, creating an optimal balance for applications requiring both English excellence and cross-lingual competency.
\begin{table}[!htbp]
\centering
\scriptsize 
\setlength{\tabcolsep}{1pt} 
\begin{tabularx}{\textwidth}{l c *{18}{>{\centering\arraybackslash}X}}
\toprule
\textbf{Models} & \textbf{Avg.} & \textbf{AR} & \textbf{BN} & \textbf{EN} & \textbf{ES} & \textbf{FA} & \textbf{FI} &	\textbf{FR} & \textbf{HI}	& \textbf{ID} & \textbf{JA} &	\textbf{KO} & \textbf{RU} & \textbf{SW} & \textbf{TE} &	\textbf{TH} & \textbf{ZH} & \textbf{DE} & \textbf{YO} \\
\midrule
\multicolumn{20}{c}{\textit{First-stage Retriever}} \\
\midrule
\jinaembeddings{} & 58.90 &	71.53 &	69.86 &	48.37 &	46.91 &	54.13 &	71.15 &	50.90 &	55.05 &	47.83 &	56.46 &	64.76 &	55.63 &	54.07 &	70.48 &	73.56 &	55.29 &	49.18 &	65.01 \\
\midrule
\multicolumn{20}{c}{\textit{Second-stage Reranker}} \\
\midrule
\model{} & 66.83 & 78.85 & 79.47 & \textbf{59.45} & 54.57 & 57.70 & 76.03 & 55.74 & 61.52 & 57.43 & 65.94 & 73.60 & 65.50 & 64.54 & 74.53 & 81.57 & \textbf{65.60} & 56.74 & 74.07 \\
\jinarerankerme{} & 66.75 & 79.78 & 78.01 & 59.21 & 53.55 & 58.90 & 70.00 & 56.66 &	62.83 &	54.92 &	66.51 &	72.86 &	67.37 &	59.04 &	70.19 &	80.37 &	64.51 &	58.50 &	80.44 \\
\jinarerankerv{} & 63.65 & 72.50 & 79.42 & 46.66 & 51.54 & 57.81 & 73.05 & 50.90 &	60.94 &	56.66 &	59.15 &	72.60 &	53.43 &	66.47 &	74.62 &	77.75 &	62.49 &	53.06 &	76.69 \\
\bgereranker{} & \textbf{69.32} & \textbf{80.51} & \textbf{81.85} & 57.67 & \textbf{57.64} & \textbf{61.92} & \textbf{80.38} &	\textbf{59.60} &	67.66 &	\textbf{58.86} &	\textbf{67.37} &	\textbf{75.14} &	\textbf{67.61} &	\textbf{68.92} &	\textbf{76.69} &	\textbf{82.29} &	64.46 &	\textbf{58.32} &	80.85 \\
\mxbaibase{} & 55.32 & 71.08 & 58.21 & 56.61 & 48.89 & 46.59 & 64.92 &	50.47 &	44.75 &	49.48 &	57.99 &	64.88 &	54.16 &	48.40 &	55.15 &	72.71 &	58.44 &	20.33 &	72.66 \\
\mxbailarge{} & 57.94 &	71.38 &	63.48 &	57.55 &	49.14 &	48.38 &	66.70 &	51.61 &	45.12 &	49.05 &	56.61 &	64.98 &	54.80 &	51.79 &	62.41 &	74.51 &	62.29 &	38.66 &	74.42 \\
\qwenrerankersmall{} & 56.16 & 67.44 & 66.67 & 50.91 & 45.77 & 52.07 & 65.50 & 43.28 & 60.36 & 49.66 & 51.56 & 61.03 & 48.88 & 46.72 & 69.86 & 72.95 & 45.14 & 43.00 & 70.04 \\
\qwenrankerlarge{} & 67.52 &	78.32 &	81.51 &	59.37 &	53.07 &	61.63 &	78.70 &	55.02 &	\textbf{68.71} &	54.90 &	65.32 &	71.80 &	64.66 &	66.50 &	75.60 &	82.00 &	59.35 &	57.56 &	\textbf{81.39} \\
\bottomrule
\end{tabularx}
\caption{Multilingual retrieval performance on the MIRACL (measured by nDCG@10).}
\label{tab:miracl_eval_result}
\end{table}

\section{Conclusion}

We present \model{}, a 0.6B-parameter multilingual listwise reranker that introduces \emph{last but not late} interaction for efficient document reranking. Our approach enables cross-document interactions during encoding by processing queries and multiple documents simultaneously within shared context windows, then extracting contextual embeddings from designated special token positions. By adapting long-context generative LLMs into a discriminative model, \model{} bridges the efficiency-effectiveness gap while maintaining significant computational advantages over generative rerankers. Future work includes studying the robustness of ranking against prompt injections and deduplication within the context window using submodularity optimization.

\bibliographystyle{plainnat}
\bibliography{main}

\appendix

\section{Model Configuration and Training Details}

\begin{table}[!htbp]
\centering
\small
\begin{tabular}{lc}
\toprule
\textbf{Parameter} & \textbf{Value} \\
\midrule
Total Parameters & 0.6B \\
Non-Embedding Parameters & 0.44B \\
Hidden Size & 1,024 \\
Number of Layers & 28 \\
Attention Heads (Q/KV) & 16/8 (GQA) \\
Context Length & 131,072 \\
Effective Sequence Length & 8,192 \\
Projector Architecture & 1024$\rightarrow$512$\rightarrow$512 \\
Projector Activation & ReLU \\
\bottomrule
\end{tabular}
\caption{Model architecture configuration for \model{}.}
\label{tab:model_config}
\end{table}

\begin{table}[!htbp]
\centering
\small
\begin{tabular}{lccc}
\toprule
\textbf{Hyperparameter} & \textbf{Stage 1} & \textbf{Stage 2} & \textbf{Stage 3} \\
\textbf{} & \textbf{Foundation} & \textbf{Context \& Hard Mining} & \textbf{Model Ensemble} \\
\midrule
Learning Rate & 5e-5 & [5e-5, 6e-6] & - \\
Batch Size (per device) & 60 & [6, 60] & - \\
Max Sequence Length & [768, 2048] & [2048, 8192] & - \\
Max Query Length & - & [256, 512] & - \\
Max Doc Length & - & [512, 2048] & - \\
Number of Negatives & 15 & [9, 25] & - \\
In-batch Negatives & 3 & [0, 3] & - \\
Temperature & 0.25 & [0.05, 0.25] & - \\
Training Mode & LoRA & LoRA/Full & Linear Merging \\
LoRA Rank & 16 & 16 & - \\
Word Embeddings & Tuned & Frozen/Tuned & - \\
Backbone & Frozen & Frozen/Tuned & - \\
Dispersive Loss $\alpha$ & 0.45 & [0.25, 0.45] & - \\
Dual Matching $\alpha$ & 0.85 & [0.65, 0.85] & - \\
Similarity Loss $\alpha$ & 0.85 & [0.75, 0.85] & - \\
\bottomrule
\end{tabular}
\caption{Multi-stage supervised fine-tuning hyperparameters showing ranges across 47 training configurations.}
\label{tab:training_config}
\end{table}

\end{document}